\documentclass[11pt]{article}
\usepackage{graphicx}
\usepackage{hyperref}
\usepackage{amsmath}
\usepackage{float}
\usepackage[percent]{overpic} 
\usepackage{geometry}
\usepackage{longtable}
\usepackage{subcaption} 
\usepackage{authblk}
\title{Advances in Transformers for Robotic Applications: A Review}
\author[1]{Nikunj Sanghai}
\author[2]{Nik Bear Brown}
\affil[1]{University of California, Los Angeles}
\affil[2]{Northeastern University}
\affil[*]{Corresponding author: Nikunj Sanghai, email: \texttt{nikunjsanghai@ucla.edu}}
\date{}

\begin{document}

\maketitle

\begin{abstract}
The introduction of Transformers architecture has brought about significant breakthroughs in Deep Learning (DL), particularly within Natural Language Processing (NLP). Since their inception, Transformers have outperformed many traditional neural network architectures due to their "self-attention" mechanism and their scalability across various applications. In this paper, we cover the use of Transformers in Robotics. We go through recent advances and trends in Transformer architectures and examine their integration into robotic perception, planning, and control for autonomous systems. Furthermore, we review past work and recent research on use of Transformers in Robotics as pre-trained foundation models and integration of Transformers with Deep Reinforcement Learning (DRL) for autonomous systems. We discuss how different Transformer variants are being adapted in robotics for reliable planning and perception, increasing human-robot interaction, long-horizon decision-making, and generalization. Finally, we address limitations and challenges, offering insight and suggestions for future research directions.
\end{abstract}

\section{Introduction}
Researchers in the domain of robotics have long relied on classical algorithms that have been proven to be extremely effective for a limited range of tasks in structured and predictable environments. Early robotic systems often relied on rule-based systems, where predefined rules could handle a limited range of tasks in a structured environment. A notable early example will be Shakey, developed by Stanford AI Lab in the 1960s which could implement logical reasoning and planning \cite{Nilsson1984}. By the 1990s, probabilistic methods started to be used with Markov Localization to account for sensor data noise and uncertainty in dynamic unstructured environment\cite{Fox_1999}. Stanford's Stanley, won the 2005 DARPA Grand Challenge, the milestone marked a shift from rule-based approaches to methods that integrated learning and probabilistic reasoning using machine learning and other data-based approaches to tackle the unpredictability of real world environments\cite{Thrun2006Stan}. However, challenges persisted in developing robots that can operate robustly in dynamic outdoor settings \cite{Mulligan2006J}. The transition from classical algorithms to probabilistic approaches has been crucial in addressing uncertainties in robot perception and action \cite{Bongard2008}. 

The increase in capabilities of Deep Learning architectures began to be integrated into autonomous systems, especially in perception. AlexNet \cite{Krizhevsky2012} outperformed all classical computer vision (CV) algorithms and popularized convolutional neural networks (CNN).  Adoption of Recurrent Neural Networks(RNNs)\cite{Rumelhart1986} and Long Short-Term Memory (LSTM) networks\cite{Hochreiter1997}, which enabled robots to process time-series data for tasks like speech recognition and navigation. Recent advances in scene understanding, sensor fusion, and multimodal approaches have further improved autonomous navigation capabilities\cite{Wijayathunga2023}. Despite these developments, achieving robust autonomy in unfamiliar outdoor environments remains a significant challenge, requiring sophisticated scene understanding and adaptive navigation strategies \cite{Glaser2012}\cite{Wijayathunga2023}.

Transformers \cite{vaswani2023} have emerged as the dominant architecture in deep learning, especially for NLP applications \cite{patwar2023},  Transformers employ self-attention  to capture long-term dependencies and model relationships across data, independent of sequence length. Their adaptability and scalability have made them powerful tools in various applications. Initially designed for NLP \cite{vaswani2023}, transformers have been introduced in robotics \cite{zhang2024hirt} \cite{zeng2023}\cite{wang2024l}, Signal Processing \cite{Verma2021AudioTT}\cite{Isik2023HPCNeuroNetAN}\cite{Chang2021EndtoEndMT}, Medical Imaging \cite{Azad2023AdvancesIM}\cite{Li2022TransformingMI}\cite{Shamshad2022TransformersIM}\cite{Henry2022VisionTI}\cite{Xiao2023TransformersIM}\cite{He2022TransformersIM}\cite{jumper2021alphafold}, and many other areas. Transformers have consistently outperformed RNNs and CNNs in various sequence modeling tasks\cite{Shin2021}. Since then, innovations such as Bidirectional Transformers for Language Understanding(BERT) \cite{Devlin2019}, Generative Pre-trained Transformer(GPT) \cite{Radford2018} , and Vision Transformers \cite{dosovitskiy2021} have extended their applicability to robotics. This paper explores the role of transformers in advancing robotic perception, control, and decision making\cite{patwar2023}. In robotics, Transformers are being adopted in three major ways: (1) as pre-trained foundation models facilitating Human-Robot Interaction and generalization\cite{Padmanabha2024}\cite{zeng2023}\cite{brohan2023rt2}\cite{brohan2023}\cite{kim2024openvla}. (2) as transformer variants integrated with Deep Reinforcement Learning enhancing long-horizon planning (3) To enhance perception, planning, and control systems. Recent models like Decision Transformer \cite{chen2021} and Trajectory Transformer \cite{janner2021} highlight how Transformers can be applied in robotics, enabling complex task planning and autonomous control in dynamic scenarios \cite{agarwal2023}\cite{li2023}. There have been many survey papers covering the use of Transformers as pre-trained foundation models in robotics \cite{zeng2023}\cite{wang2024l}\cite{Jeong2024l}\cite{Firoozi2023l}\cite{hu2024general}\cite{Kim_2024}\cite{Kawaharazuka2024}\cite{Xiao2023TransformersIM}. Some survey papers have covered integration of transformers in RL covering all applications \cite{agarwal2023}\cite{li2023}\cite{Hu2024survey} , while some have exclusively focused on application in Robotics\cite{moroncelli2024}\cite{Yuan_2023}.Our paper gives a broad overview of all the different ways Transformers have been used in Robotics. At the same time, the scope of our paper is narrower than surveys covering various advances in Transformers \cite{Tay2020E}\cite{Islam2023ACS}\cite{xu2023}\cite{LIN2022}, in this paper we wish to exclusively focus on applications in robotics. Some Transformer architectures have been tailored for robotic perception, planning, and control, other areas of usage include pre-trained foundation models  to specialized variants used in DRL. We also address current limitations and propose directions for future research with the goal of enhancing adaptability, generalization, and long-horizon decision making in robotic systems.
\section{Background}
This section introduces the fundamental concepts of the original Transformers \cite{vaswani2023} architecture. We start with an overview and then discuss the core components.  
\subsection{Overview of Transformer Architecture}
Traditional sequence models like RNNs were capable of handling variable-length inputs and temporal dependencies \cite{DiPietro2020} \cite{Peng2009}, LSTMs were a variant of RNNs with an added gating mechanism that addressed long-term dependency problems \cite{Yu2019} \cite{Zhang2023} to an extent, but due to the inherent way RNNs and LSTMs process data, each time step depends on the previous step. This dependency limits parallel processing and struggles with long-term dependencies, as information from earlier inputs can fade over long sequences due to issues such as vanishing gradients \cite{Le2016}\cite{AlSelwi2023} and explosion \cite{Sherstinsky2018}. CNNs, though useful for capturing local dependencies in sequences, lack the ability to model long-range dependencies.

The Transformer model addresses these limitations by employing self-attention and eliminating recurrence and convolution. Self-attention, also referred to as intra-attention, connects various positions within a single sequence to generate a meaningful representation of that sequence\cite{vaswani2023}. Self-attention allows each input position to focus on relevant positions across the entire sequence simultaneously, which not only captures dependencies at varying distances but also enables efficient parallelization during training and inference
Transformer-based models like BERT and GPT have achieved state-of-the-art results in machine translation and other language tasks \cite{Zhu2024}\cite{Sutskever2014}.
\subsection{Core Components of Transformer Architecture}
The Transformer architecture consists of an encoder-decoder architecture\cite{vaswani2023}, each consisting of multiple layers stacked together, with each layer containing sublayers designed to carry out a particular operation. The encoder in sequence transduction models processes an input sequence of symbolic representations $(x_1, \ldots, x_n)$ and transforms it into a corresponding sequence of continuous representations $\mathbf{z} = (z_1, \ldots, z_n)$. The decoder then uses $\mathbf{z}$ to generate an output sequence $(y_1, \ldots, y_m)$, producing one symbol at a time. At each generation step, the model operates in an auto-regressive manner, relying on both the encoded representations and the previously generated outputs as additional inputs for predicting the next symbol.

\begin{figure}[htbp]
    \centering
    \includegraphics[width=1.0\textwidth]{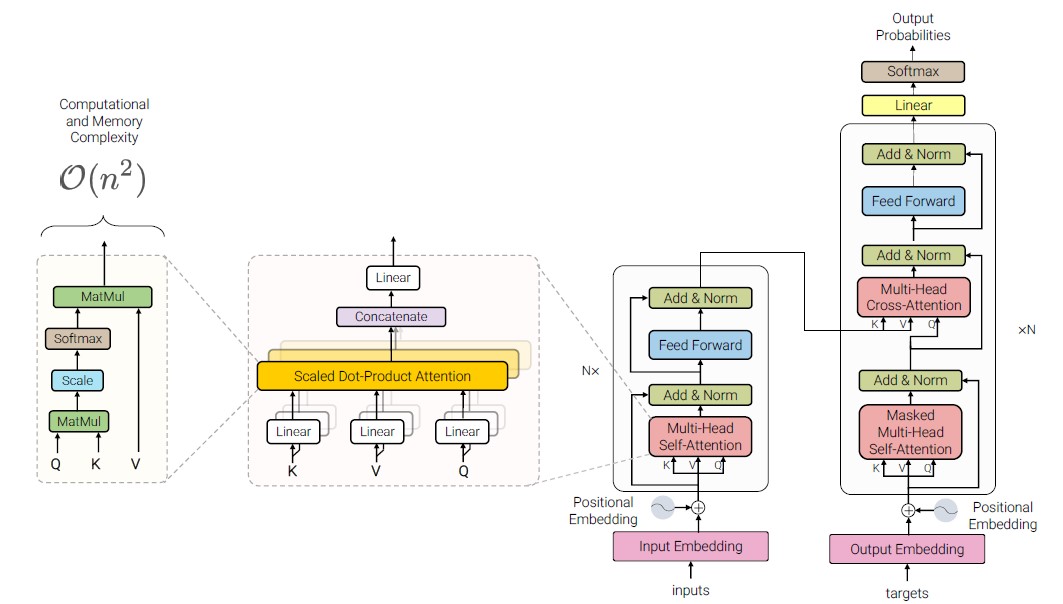} 
    \caption{The Original "Vanilla" Transformer Architecture \cite{vaswani2023}. Visual representation taken from\cite{Tay2020E}}
    \label{fig:transformer}
\end{figure}

\subsubsection{Encoder and Decoder Architecture}

\textbf{Encoder}: The input sequence is tokenized into smaller units, such as characters or words. The encoder converts the input sequence into a series of contextual embeddings that capture the relationship between these tokens in the input sequence. Each encoder layer has two main components: 1)Self-Attention Mechanism: It enables each token in the input sequence to consider and relate to other tokens, capturing relations between the sequence 2)Feed-Forward Neural Network: The neural network applies non-linear transformations independent of the embeddings of each token. To improve training stability and address issues like vanishing gradient, residual connections are used around each sub-layer. These connections add the input of a sub-layer to its output, followed by layer normalization. The output of each sub-layer is 
\[
\text{LayerNorm}(\mathbf{x} + \text{Sublayer}(\mathbf{x})),
\]
where \(\text{Sublayer}(\mathbf{x})\) is the output of the sub-layer applied to the input x. 
\\
\\
\textbf{Decoder}: The decoder generates the output sequence by combining the encoded representation from the encoder along with the previously generated outputs to predict the next token. Each decoder layer has three sub-layers: 1) Masked Self-Attention Mechanism: Similar to the encoder's self-attention, this mechanism ensures each token can only attend to previous tokens and itself, preventing any dependence on future tokens. 2) Encoder-Decoder Attention: This sub-layer uses multi-head attention to allow the decoder to focus on relevant parts of the encoder's output, helping it incorporate meaningful input context while generating the sequence 3) Feed-Forward Neural Network: A fully connected layer independently processes each token to learn more complex patterns in the sequence.

As in the encoder, residual connections are applied around each sub-layer, and layer normalization is performed. This ensures smoother gradient flow and better convergence during training. The decoder combines these mechanisms to generate coherent output token-by-token.

\subsubsection{Self-Attention Mechanism}

Self-attention allows each token to be considered and related to every other token in the sequence, enabling the model to learn relationships across long distances.

\textbf{Scaled Dot-Product Attention}: Self-attention is implemented using a scaled dot-product mechanism, where three vectors are computed for each input token:

\begin{itemize}
    \item \textbf{Key (K)}: Represents the token holding relevant information.
    \item \textbf{Value (V)}: Contains the actual information that is distributed based on the
    \item \textbf{Query (Q)}: Represents the token making a request for related information. attention scores.
\end{itemize}

The attention score between a query \( q \) and a key \( k \) is calculated as:

\[
\text{Attention}(Q, K, V) = \text{softmax} \left( \frac{QK^T}{\sqrt{d_k}} \right) V
\]

Here, \( d_k \) is the dimensionality of the keys, and the score is scaled by \( \sqrt{d_k} \) to prevent exceedingly large values that could affect the gradient flow.

\textbf{Multi-Head Attention}: Transformer architecture uses multiple attention heads to capture different aspects of the relationships in the same sequence. Each head has its own learned linear transformations, allowing the model to learn different types of dependencies across the sequence.

Multi-head attention is calculated by concatenating the outputs of multiple attention heads followed by a linear transformation:

\[
\text{MultiHead}(Q, K, V) = \text{Concat}(\text{head}_1, \ldots, \text{head}_h) W^O
\]

where \( W^O \) is a learned weight matrix for the output.

\subsubsection{Positional Encoding}

Since the Transformer lacks recurrence and does not inherently process tokens sequentially, it must encode positional information to understand the order of tokens. The positional encoding vector is added to the input embeddings to provide each position with unique positional information.

The positional encoding for each position \( pos \) and dimension \( i \) is given by:

\[
\text{PE}(pos, 2i) = \sin \left( \frac{pos}{10000^{2i/d_{model}}} \right)
\]

\[
\text{PE}(pos, 2i+1) = \cos \left( \frac{pos}{10000^{2i/d_{model}}} \right)
\]

where \( d_{model} \) is the dimension of the embeddings. These sine and cosine functions encode relative positions in a way that is interpretable by the model.

\subsubsection{Feed-Forward Neural Network (FFN)}

Each position in the sequence is independently passed through a fully connected feed-forward network. The FFN consists of two linear transformations with a ReLU activation in between:

\[
\text{FFN}(x) = \text{max}(0, x W_1 + b_1) W_2 + b_2
\]

where \( W_1 \), \( W_2 \), \( b_1 \), and \( b_2 \) are learnable parameters. This component allows for non-linear transformation, which improves the model's ability to capture complex relationships.

\subsection{Architectural Innovations in Transformers}
Transformers have undergone numerous enhancements to address limitations of the architecture,  the quadratic computational complexity of the attention mechanism that limits their ability to handle long sequences \cite{Ziyaden2021L} being just one of them. We will briefly go over some prominent innovations.  

\subsubsection{Efficient Transformers}
Efficient transformer variants try to address the computational challenges of long sequences by reducing the quadratic time complexity of the original attention mechanism, making transformers feasible for resource-constrained or real-time environments. There are many different approaches researchers have taken, Performer \cite{choromanski2022r} replaces softmax attention mechanism with kernel-based approximation function, thus being able to handle much longer sequences compared to original transformer, and drastically reduces the memory footprint along with linear computational complexity. Linformer \cite{wang2020lin} reduces the attention complexity to linear time by projecting the attention matrix, allowing for efficient processing of large state-action spaces. 

\begin{longtable}{|p{4cm}|p{2cm}|p{9cm}|}
\hline
\textbf{Model} & \textbf{Complexity} & \textbf{Key Technique} \\
\hline
\endfirsthead
\hline
\textbf{Model} & \textbf{Complexity} & \textbf{Key Technique} \\
\hline
\endhead
Transformer \cite{vaswani2023} & $O(n^2)$ & Original transformer architecture with quadratic time complexity. \\
\hline
Performer \cite{choromanski2022r} & $O(n)$ & Kernel-based approximation for softmax attention; \\
\hline
Linformer \cite{wang2020lin} & $O(n)$ & Projection of attention matrix into lower-dimensional space; linear complexity. \\
\hline
Synthesizer \cite{tay2020sparse} & $O(n)$ & Learned or random synthetic attention for reduced complexity. \\
\hline
Nyströmformer \cite{xiong2021nystrom} & $O(n)$ & Nyström method for approximating self-attention; \\
\hline
Linear Transformer \cite{katharopoulos2020} & $O(n)$ & Kernel-based approximation of softmax attention;  \\
\hline
FlashAttention \cite{dao2022} & $O(n^2)$ & Kernel-level optimized attention for speed and memory efficiency achieves significant runtime improvements. \\
\hline
FNet \cite{leethorp2022} & $O(n)$ & Replaces attention with Fast Fourier Transform for efficiency. \\
\hline
Hyena \cite{poli2023hyena} & $O(n \log n)$ & Lightweight convolutional structures for long-sequence processing. \\
\hline
\end{longtable}

\subsubsection{Multimodal Transformers}
Original Transformers were developed for NLP application specifically to use encoder-decoder mechanism for for text. Multimodal Transformers are capable of processing multiple type of inputs, ViLBERT \cite{lu2019vilbert} extended the popular BERT \cite{Devlin2019} architecture introducing a co-attention mechanism, by exchanging key-value pairs in multi-headed attention, the structure incorporates linguistic queues to visual ones and vice versa, see Figure~\ref{fig:multimodal} 

\begin{longtable}{|p{5cm}|p{9cm}|}
\hline
\textbf{Model} & \textbf{Key Technique} \\
\hline
\endfirsthead
\hline
\textbf{Model} & \textbf{Key Technique} \\
\hline
\endhead
CLIP \cite{radford2021} & Jointly learns embeddings for text and images, leveraging contrastive learning. \\
\hline
FLAVA \cite{singh2022flava} & Unified architecture for vision, language, and their joint representation. \\
\hline
Data2Vec \cite{baevski2022data} & Multimodal framework using the same architecture for vision, audio, and text. \\
\hline
VisualGPT \cite{li2019visualbert} & Combines image features with text generation for vision-to-language tasks. \\
\hline
DALL-E \cite{ramesh2021zero} & Text-to-image generation using multimodal Transformer architecture. \\
\hline
ALIGN  \cite{jia2021scaling} & Pretrained on large-scale image-text pairs for robust visual and language tasks. \\
\hline
MERLOT \cite{zellers2021} & Learns multimodal representation for video and text understanding. \\
\hline
VideoBERT \cite{sun2019video} & Processes video frames and associated text for video-language tasks. \\
\hline
\end{longtable}

\begin{itemize}
    \item \textbf{VisualBERT} \cite{li2019visualbert}: Integrates text and image data, with potential applications in robotic vision systems where contextual understanding of visual inputs is required. For example, in human-robot interaction scenarios, understanding both spoken commands and visual context enhances a robot's ability to interact with humans naturally.
    \item \textbf{UNITER} \cite{li2020oscar}: Pretrained on multimodal datasets, it is useful for tasks requiring joint text and image understanding, such as scene description and semantic mapping, which can improve the situational awareness of service robots.
\end{itemize}

\subsubsection{Sparse and Adaptive Transformers}
Sparse transformers introduce sparsity in the self-attention mechanism to reduce computational overhead without sacrificing performance. Adaptive transformers dynamically adjust computation based on input data characteristics, optimizing resource usage—crucial for deploying transformers in real-time robotic systems. Longformer \cite{beltagy2020long} introduces a combination of global and sliding window local attention mechanisms, suitable for tasks requiring long-range dependencies.  Reformer \cite{kitaev2020reform} Reduces time complexity in processing large datasets through locality-sensitive hashing.

\begin{longtable}{|p{4cm}|p{2cm}|p{9cm}|}
\hline
\textbf{Model} & \textbf{Complexity} & \textbf{Key Technique} \\
\hline
\endfirsthead
\hline
\textbf{Model} & \textbf{Complexity} & \textbf{Key Technique} \\
\hline
\endhead
Sparse Transformer\cite{child2019} & $O(n \sqrt{n})$ & Sparse attention with structured sparsity patterns for efficient long-sequence modeling. \\
\hline
Longformer \cite{beltagy2020long}& $O(n)$ & Global and sliding window local attention, suited for long-range dependencies. \\
\hline
BigBird \cite{zaheer2021} & $O(n)$ & Combination of random, sliding window and global attention mechanisms. \\
\hline
Routing Transformer \cite{roy2020} & $O(n \log n)$ & Sparse attention using K-means clustering-based routing. \\
\hline
Reformer \cite{kitaev2020reform} & $O(n \log n)$ & Locality-sensitive hashing (LSH) for reduced time complexity. \\
\hline
Routing Transformer \cite{roy2020} & $O(n^{1.5}d^*)$ & K-means clustering-based sparse attention routing. \\
\hline
Sinkhorn Transformer\cite{tay2020sparse} & $O(n)$ & Differentiable sorting in attention for structured inputs. \\
\hline
Mega \cite{ma2023mega} & $O(n)$ & Combines sparse and memory-efficient attention for general long-sequence modeling. \\
\hline
\end{longtable}
\noindent
\textit{*Here, $d$ represents the hidden dimension of the model.}
\begin{itemize}
    \item \textbf{Adaptive Attention Span Transformer} \cite{sukhbaatar2019adaptive}: Adjusts the attention span dynamically extends significantly the maximum context size used in Transformer, while maintaining control over their memory footprint and computational time. Many researchers have considered different approaches across various applications some names include Universal Transformer \cite{dehghani2019universal}, Dynamic Vision Transformer \cite{wang2021images} and  Feedback Transformer \cite{fan2021addres}. 
\end{itemize}
\subsection{Overview of Reinforcement Learning}

Reinforcement Learning (RL) is a computational approach of understanding and automating goal-directed learning and decision-making. In RL, an agent learns to make decisions by performing actions in an environment to maximize cumulative rewards \cite{sutton2018reinforcement}. The agent interacts with the environment in discrete time steps, receiving observations and rewards, and updates its policy—a mapping from states to actions—based on this feedback.

In the context of robotics, RL provides a policy for robots to learn optimal actions through trial and error, without explicit programming of the desired task \cite{kober2013reinforcement}. RL algorithms can be categorized into model-based and model-free methods. Model-based RL involves learning a model of the environment's dynamics, while model-free RL directly learns the policy or value function without an explicit model.

Classical RL approaches faced challenges in handling high-dimensional state and action spaces typical in robotics. However, the integration of deep learning with RL, known as Deep Reinforcement Learning (DRL), has enabled the handling of complex, high-dimensional inputs like images and sensor data \cite{mnih2015human}. DRL has shown success in various domains, including robotic manipulation, navigation, and control tasks \cite{levine2016end}.

\subsection{Innovations in Reinforcement Learning}

Recent innovations in RL focus on improving sample efficiency, stability, and generalization of learning algorithms. Key advancements include:

\begin{itemize}
    \item \textbf{Policy Gradient Methods}: Algorithms like Proximal Policy Optimization (PPO) \cite{schulman2017proximal} and Trust Region Policy Optimization (TRPO) \cite{schulman2017trust} have improved the stability of policy updates, facilitating more reliable training in continuous action spaces common in robotics.

    \item \textbf{Off-Policy Learning}: Techniques such as Deep Deterministic Policy Gradient (DDPG) \cite{lillicrap2019conti} and Soft Actor-Critic (SAC) \cite{haarnoja2018soft} enable efficient learning from off-policy data, enhancing sample efficiency which is critical when interactions with the real world are costly or limited.

    \item \textbf{Model-Based RL}: Methods that incorporate learning a model of the environment, such as Model-Based Policy Optimization (MBPO) \cite{janner2021trust}, have shown improved sample efficiency by utilizing simulated experiences, which is particularly beneficial in robotics where real-world data collection is expensive.

    \item \textbf{Meta-RL and Transfer Learning}: Approaches like Model-Agnostic Meta-Learning (MAML) \cite{finn2017modelagnostic} allow agents to rapidly adapt to new tasks with minimal data, addressing the challenge of generalization in RL.

    \item \textbf{Hierarchical Reinforcement Learning (HRL)}: Techniques such as the Options Framework \cite{bacon2016opt} and Hierarchical DDPG \cite{nachum2018dataeff} enable agents to learn policies at multiple levels of abstraction, making them efficient in solving tasks with hierarchical structures, such as multi-step navigation or manipulation \cite{bacon2016opt}.

    \item \textbf{Exploration Strategies}: Strategies like intrinsic motivation (e.g., curiosity-driven learning) \cite{pathak2017curiositydriven} and uncertainty-aware exploration (e.g., bootstrapped DQN) \cite{osband2016deepexpl} enhance RL's ability to explore efficiently in sparse-reward environments.

    \item \textbf{Safety in RL}: Methods such as Constrained Policy Optimization (CPO) \cite{achiam2017} and Reward Shaping \cite{ng1999reward} introduce safety constraints during training, ensuring safe exploration and operation, especially important in robotics.

    \item \textbf{Multi-Agent Reinforcement Learning (MARL)}: Frameworks like Multi-Agent Deep Deterministic Policy Gradient (MADDPG) \cite{lowe2020multiagent} and QMIX \cite{rashid2018} extend RL to multi-agent settings, enabling collaboration and competition in complex environments like swarm robotics or team-based games.

    \item \textbf{Offline Reinforcement Learning}: Techniques like Conservative Q-Learning (CQL) \cite{kumar2020} train agents on previously collected datasets without additional interaction with the environment, making them practical for real-world scenarios where data collection is expensive.

    \item \textbf{Incorporation of Attention Mechanisms}: The integration of attention mechanisms, particularly through Transformer architectures, has enhanced the ability of RL agents to focus on relevant parts of the input space, improving performance in tasks requiring long-term dependencies and memory \cite{parisotto2019stabilizing}.
\end{itemize}

These innovations have significantly advanced the capability of RL algorithms, making them more applicable to the complex and dynamic environments encountered in robotics.

\section{Applications in Robotics}

Recently there have been numerous studies that have applied Transformers architecture, we will be broadly discuss various applications in detail. 

\subsection{Transformers as Pre-trained Foundation Models}
In this section we discuss primarily 2 broad topics in which foundation models are being utilized in zero-shot robot generalization and Human-Robot collaborations. Pretrained Large Language Models (LLMs) \cite{Radford2018}, Large Vision-Language Models (VLMs), and Vision-Language-Action Models (VLA) \cite{brohan2023}\cite{brohan2023rt2} are currently being used to increase generalization in Robotics. Prior to the emergence of foundation models, traditional deep learning models for robotics were typically trained on limited datasets gathered for distinct tasks \cite{Sun_2022} \cite{Firoozi2023l}. Much progress has been done via collaboration, datasets such as Open X-Embodiment \cite{embodimentcollaboration2024} consist of data from 22 different robot embodiments, demonstrating 527 skills over 160,266 tasks,  containing 1M+ real robot trajectories. The dataset was created by pooling 60 existing datasets from 34 research labs. The datasets were preprocessed into a consistent data format RLDS \cite{ramos2021rlds} for better access. 

\begin{figure}[H]
    \centering
    \begin{subfigure}[b]{0.4\textwidth}
        \includegraphics[width=\textwidth]{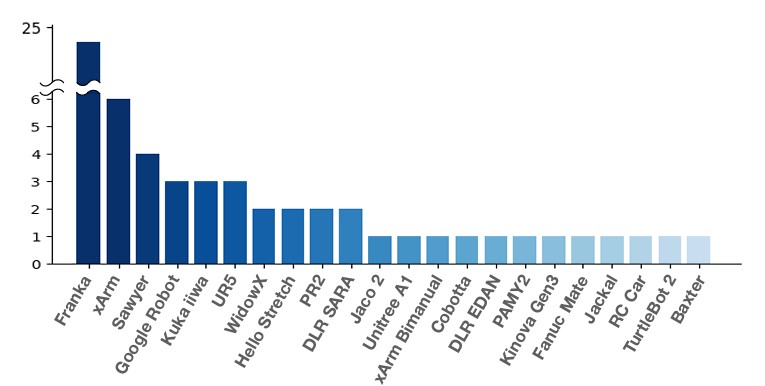}
        \caption{Number of Robot Datasets}
        \label{fig:datasets_per_robot.jpg}
    \end{subfigure}
    \hfill
    \begin{subfigure}[b]{0.31\textwidth}
        \includegraphics[width=\textwidth]{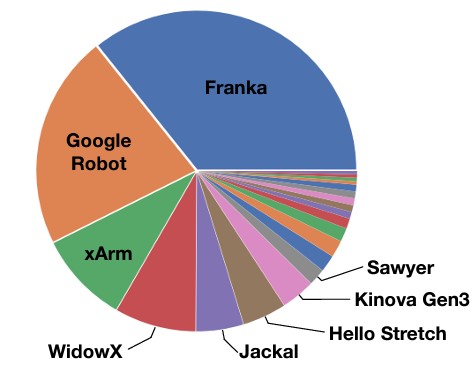}
        \caption{Scenes per Robot }
        \label{fig:scenes_for_emb.jpg}
    \end{subfigure}
    \hfill
    \begin{subfigure}[b]{0.25\textwidth}
        \includegraphics[width=\textwidth]{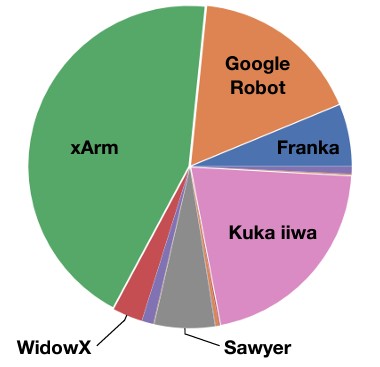}
        \caption{Trajectories per Robot}
        \label{fig:trajectories_for_emb.jpg}
    \end{subfigure}

    \vspace{0.5cm} 

    \begin{subfigure}[b]{0.4\textwidth}
        \includegraphics[width=\textwidth]{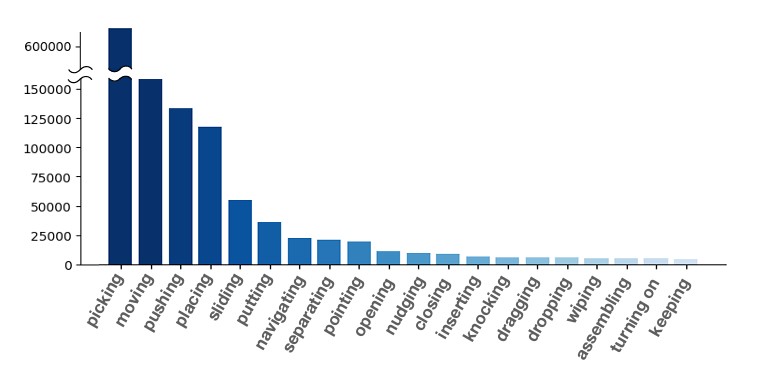}
        \caption{Common Dataset skills}
        \label{fig:common_dataset_skills.jpg}
    \end{subfigure}
    \hfill
    \begin{subfigure}[b]{0.59\textwidth}
        \includegraphics[width=\textwidth]{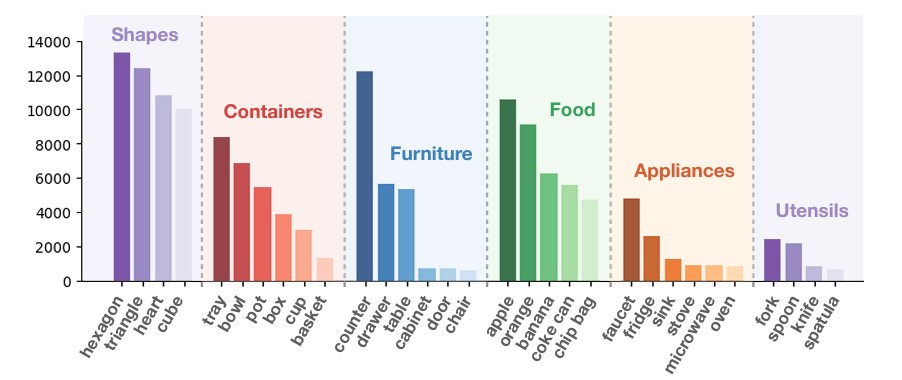}
        \caption{Common Dataset Objects}
        \label{fig:common_dataset_objects.jpg}
    \end{subfigure}

    \caption{ The Open X-Embodiment(OXE) Dataset, graphs are from the original paper. a) The dataset consists of 60 datasets across 22 embodiments. The dataset has a wide range of skills and object interactions please see  ~\ref{fig:common_dataset_skills.jpg} and ~\ref{fig:common_dataset_objects.jpg}. While Franka has the most number of scenes see ~\ref{fig:datasets_per_robot.jpg} x-Arm and Google Robot have the biggest contribution to trajectory data see ~\ref{fig:scenes_for_emb.jpg} and ~\ref{fig:trajectories_for_emb.jpg}    \cite{embodimentcollaboration2024}}
    \label{fig:OXE_dataset}
\end{figure}
In recent times, many generalist robot models have been introduced, Gato \cite{reed2022generalist}, BC-Z\cite{jang2022bcz}, Octo \cite{octomodelteam2024}, RT-1 \cite{brohan2023}, RT-2 \cite{brohan2023rt2}, OpenVLA \cite{kim2024openvla} and $\pi_0$ \cite{black2024pi0}. In the case of RT-1 the authors have collected their own data. Most of the other models have used a subset of the OXE dataset to train their models. $\pi_0$ has taken a subset of the OXE dataset along with some proprietary closed-source data to train their model. Octo trains a generalist policy that can control multiple robots within the OXE dataset for zero-shot tasks and allows for fine-tuning to new robot setup. The performance of Octo fluctuates depending on the robot used.Prior works like Octo typically compose pretrained components such as t5-base \cite{colin2020} for tokenization of language instructions with additional original subsystems of the architecture that are fine-tuned for performance benefits during training.OpenVLA's approach can be characterized as an end-to-end pipeline, developed by directly fine-tuning VLMs to generate robot actions see Figure ~\ref{comparison_SOTA}.  RT-2-X  trains a 55B-parameter VLA policy on the Open X-Embodiment dataset, but the model is limited to cloud deployment because of the size of the model, in comparison models like Octo are much smaller 100M parameters. OpenVLA is able to outperform RT-2-X models on a much smaller memory footprint, further quantization done by authors shows no signs of significant drop in performance. While RT-1, RT-2, and $\pi_0$ are close-source, Octo and OpenVLA are open-source models. 
\begin{figure}[H]
    \centering
    \begin{subfigure}[b]{0.87\textwidth}
        \includegraphics[width=\textwidth]{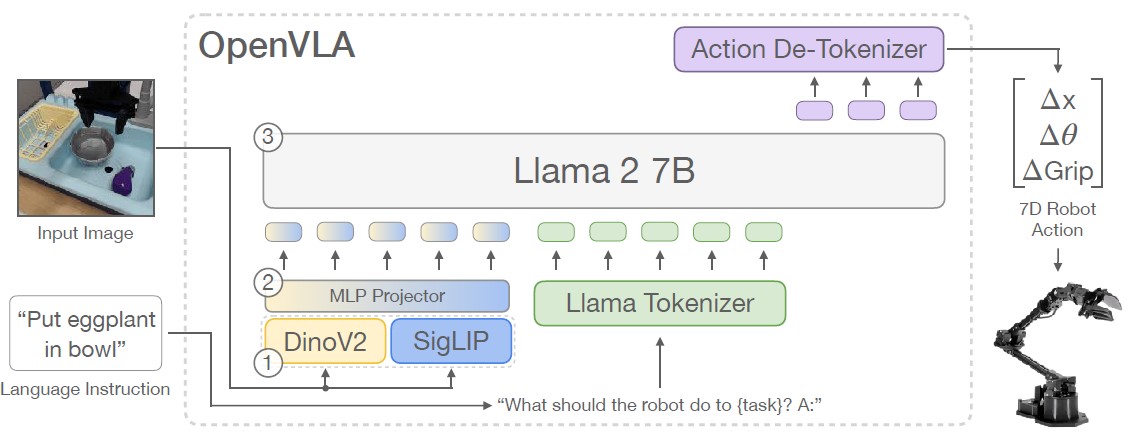}
        \caption{OpenVLA architecture }
    \end{subfigure}
    \caption{ Visual representation of OpenVLA architecture, generating low-level robot action from high-level human commands. Image representations taken from original authors\cite{kim2024openvla}.}
    \label{fig:comparison_SOTA}
\end{figure}

Transformers have been utilized as pre-trained foundation models to improve human-robot interaction (HRI) and human-robot collaboration (HRC), giving robots the ability to discern tasks from human speech \cite{Padmanabha2024}\cite{Park_2024} \cite{park2024integrat}. Using large-scale language models such as BERT \cite{Devlin2019} and GPT \cite{Radford2018}, robots can better understand and process natural language commands, leading to more intuitive and effective communication with humans, even taking personal preferences\cite{newman2024degust}\cite{Newman2024ho}\cite{wang2024apricot} into account. It is to be noted that human commands can introduce ambiguity and uncertainty in robot actions \cite{huang2022lan}. Some newer research has shown the use of human feedback to correct errors in planning \cite{burns2024plan}\cite{zha2024dis}.

 Models like \textbf{LAMA} \cite{petroni2019} demonstrate the potential of pre-trained Transformers in capturing factual knowledge, which can be applied in robots for question-answering and dialogue systems. Additionally, integrating multimodal Transformers that process both language and visual inputs enhances a robot's ability to interpret instructions within the context of its environment \cite{lu2019vilbert}. To improve trust, there has also been research to evaluate policy for communication from autonomous robots to humans, to build trust with humans \cite{kim2023explain}.

In HRI, pre-trained Transformers enable:

\begin{itemize} \item \textbf{Natural Language Understanding}: Improved parsing and comprehension of human language, allowing robots to interpret and execute complex instructions \cite{Tellex2020RobotsTU}.

\item \textbf{Contextual Awareness}: Ability to consider contextual cues from the environment and prior interactions, leading to more appropriate responses \cite{thomason2019}.

\item \textbf{Adaptability}: Enhanced learning from limited data through transfer learning, reducing the need for extensive task-specific training \cite{brown2020lang}.
\end{itemize}

Some successful case studies have demonstrated the ability of foundation models to augment Human-Robot Collaboration(HRC), researchers have demonstrated robustness with a 92.73\% detection of fault instructions by humans, and validated real-world performance in construction settings where multi-modal inputs were given to robots \cite{park2024integrat}. Validation studies to demonstrate robot performance from natural language commands from non-experts have demonstrated 99.95\% success rate in simulation in a construction setting \cite{Park_2024}. Other applications include grasping\cite{xu2024jointmodel}\cite{li2024shape}, autonomous driving\cite{wang2023drive}, assistive feeding\cite{Padmanabha2024}\cite{jenamani2024flair} and household robots \cite{Wu_2023}.

\subsection{Transformer Variants Integrated in Deep Reinforcement Learning}
Many studies have been conducted to integrate transformers to address the challenges in Deep Reinforcement Leaning(DRL) such as sample efficiency and partial observability \cite{agarwal2023}\cite{li2023}. Transformers have shown to be effective as world models, enabling sample-efficient learning in complex environments \cite{parisotto2021marine}\cite{Micheli2022Transworld}. Actor-Learner distillation has been proposed to allow a large complex "learner" model based on transformer architecture to transfer its learning progress to a smaller simpler LSTM "actor" model, having low inference times while maintaining low sample effeciency of transformers\cite{Parisotto2021Eff}. Transformers have also been used to encode temporal logic into latent task features, allowing hierarchal reinforcement learning for long-horizon manipulation \cite{Lotfi2024TRAPs}. Innovations such as the Online Decision Transformer (ODT) integrate offline pre-training with online fine-tuning, using sequence modeling for reinforcement leaning (RL), bridging the performance gap when offline RL models are suboptimal for tasks involving online interactions \cite{zheng22c}. 
Constrained Decision Transformer(CDT) integrates transformers to solve offline RL problems by learning high-reward policies while adhering to strict safety constraints \cite{liu23m}

Decision Transducer (DTd), a multimodal Transformer framework for offline reinforcement learning (RL) \cite{wang23d}. Traditional RL approaches treat trajectories (state, action, reward) as a single sequence. DTd disentangles these into three unimodal sequences (states, actions, rewards) and processes them independently before selectively fusing them for decision-making.
Transformers have been integrated with Deep Reinforcement Learning to address challenges in planning and decision-making over long horizons. The sequential modeling capabilities of Transformers enable RL agents to consider extended sequences of actions and states, improving performance in complex tasks. Language Models have also been shown to implement policy iteration faster than both imitation learning and gradient descent based RL \cite{brooks2023largelang}
Notable Transformer-based RL models include:

\begin{itemize} \item \textbf{Decision Transformer} \cite{chen2021}: This model formulates RL as a sequence modeling problem, using a Transformer to predict future actions based on past states, actions, and returns. It has shown promise in handling long-horizon tasks without explicit policy optimization.

\item \textbf{Trajectory Transformer} \cite{janner2021}: This approach models the distribution of trajectories in a dataset, enabling planning by sampling and optimizing over possible future trajectories. It allows for efficient utilization of offline datasets, beneficial in robotics where collecting new data can be expensive.

\item \textbf{GTrXL (Gated Transformer-XL)} \cite{parisotto2019stabilizing}: Enhances the stability of Transformer architectures in RL by incorporating gating mechanisms, improving performance on tasks requiring long-term memory and reasoning.

\item \textbf{Perceiver} \cite{jaegle2021perceiver}: A model capable of handling multimodal inputs and scalable to high-dimensional data, suitable for robotic applications involving complex sensory information processing.

\item \textbf{Behavior Transformers} \cite{shafiullah2022}: Combines Transformers with behavior cloning to enable RL agents to mimic human actions effectively in imitation learning setups.
\end{itemize}

The integration of Transformers in DRL offers several advantages for robotics:

\begin{itemize} \item \textbf{Long-Horizon Planning}: Ability to model dependencies over long time scales, essential for tasks like navigation and manipulation where early decisions affect long-term outcomes \cite{wu2021greedy}.

\item \textbf{Sample Efficiency}: Improved utilization of collected data through better generalization and representation learning, reducing the need for extensive interactions with the environment \cite{kostrikov2021}.

\item \textbf{Handling High-Dimensional Inputs}: Effective processing of complex sensory inputs, including images, point clouds, and proprioceptive data, enabling richer perception capabilities \cite{reed2022generalist}.

\item \textbf{Flexibility in Policy Representation}: Transformers can represent complex policies that consider sequences of past observations and actions, facilitating more sophisticated decision-making strategies \cite{furuta2022generalize}.

\item \textbf{Few-Shot and Zero-Shot Learning}: Pre-trained Transformers demonstrate robust generalization capabilities, enabling RL agents to adapt to new tasks with minimal or no additional training \cite{brown2020lang}.

\item \textbf{Scalability Across Modalities}: Transformers are inherently capable of handling multimodal data, making them ideal for robotic tasks requiring integration of vision, language, and proprioception \cite{jaegle2021perceiver}
\end{itemize}

These Transformer variants represent a significant step toward more capable and autonomous robotic systems, capable of operating in unstructured and dynamic environments.
MarineFormer \cite{parisotto2021marine} used spatio-temporal graph where the nodes included the Unmanned Surface Vehicle (USV), static and dynamic obstacles, and current flow dynamics, it demonstrated 20\% success rate improvement over state-of-the-art reinforcement learning-based methods for marine environments. Researchers have used VLA models to enhance RL capabilities by requiring robot arms to grasp objects in a cluttered environment \cite{xu2024jointmodel}. Multimodal Tranformers have been used in behavior cloning RL, InstructRL, trains a policy based on natural language and visual observations \cite{liu2023instruction}.
\subsection{Transformers in Perception, Planning and Control}
This section gives the readers a section-wise overview of current use of Transformers for perception, planning, and control. 
\subsubsection{ Perception}
    Vision Transformers (ViTs) \cite{dosovitskiy2021} have been used in Robotics, particularly in manipulation and navigation tasks. ViTs  have shown promising results in 3D object manipulation \cite{Goyal2023RVTRV} and real-world control of manipulators \cite{brohan2023}\cite{reed2022generalist}\cite{kim2024openvla}. Models like CLIP \cite{radford2021} (Contrastive Language–Image Pretraining) developed by OpenAI, have enabled robots to align textual and visual modalities effectively, allowing them to interpret human instructions, classify objects, and even adapt to novel scenarios with zero-shot learning. This capability is particularly useful for building generalizable models in unstructured environments where predefined object categories or explicit annotations may not be available. CLIP has been utilized for tasks such as real-time object recognition, semantic scene understanding, and human-robot interaction, enhancing robots' contextual awareness and adaptability. Subsequent work has decreased the memory footprint of CLIP \cite{sun2023evaclip}, expanded the applications to 3D point cloud scene understanding \cite{zhang2021point}\cite{chen2023clip} and grounding, which is the process of associating objects with text descriptions \cite{li2022grounded}\cite{rasheed2024glamm}. Meanwhile, the Segment Anything Model (SAM)\cite{kirillov2023seg}, a foundation model for image segmentation, has revolutionized how robots process and segment visual data in real-time. By generating accurate object masks from any point prompt within an image, SAM enables precise object manipulation and navigation in cluttered environments, which is critical for domains like warehouse automation and medical robotics. Researchers have integrated SAM into robotic systems for tasks like autonomous grasping and visual servoing, where the ability to identify and segment objects dynamically ensures better control and decision making. Current research has broadened the scope of the application by reducing computational cost \cite{ke2023seg} and introducing video segmentation using a memory mechanism to process frames sequentially storing and using information from previous frames to predict masks over time \cite{ravi2024sam2}. While VLMs excel at tasks like prediction or segmentation, they lack a deep understanding of the physical attributes of the objects, PG-InstructBLIP \cite{gao2024physical}, fine tunes InstructBLIP \cite{dai2023instructblip} model to improve model's ability to reason about physical concepts. 
\subsubsection{ Planning}

Transformers model sequences of actions and observations, aiding in long-horizon decision-making for autonomous robots. Their ability to maintain long-term dependencies allows them to plan paths that maximize safety and efficiency. Decision Transformers \cite{chen2021} treat RL as a sequence modeling problem, making them suitable for long-term planning in autonomous vehicles. Gated Transformer-XL \cite{dai2019transformerx} for Long-Term Memory can also be used for the same purpose. To increase robustness in planning over long horizons, researchers have proposed introducing human-like reasoning capabilities to generate sub-tasks, and getting top-k hypothesis and the best selection is made based on feedback from the environment \cite{logeswaran2022fews}. Another approach taken is to facilitate planning by creating out-of-distribution (OOD) scenarios using LLMs for both environmental scenarios such as extreme weather or road obstructions and unpredictable behavior from other road users \cite{aasi2024gen}. Often a limitation for planning could be the limited number of tasks a planner can execute, using LLMs to facilitate a natural language task breakdown, and using neural descriptor fields (NDFS) to learn new tasks in a few shots \cite{parakh2023li}. Recently, the ALOHA 2 platform uses a transformer-based architecture with diffusion policy to perform challenging bi-manual manipulation tasks \cite{zhao2024aloha}. ViNT, a Transformer-based model for visual navigation, demonstrates significant potential in long-horizon decision-making by enabling zero-shot generalization across diverse environments. It employs an EfficientNet encoder for tokenizing image inputs and a diffusion model for proposing subgoals, making it effective in complex, unstructured scenarios. Some researchers have also explored use of human feedback to improve long-horizon task success rates, and have noticed a 20\% increase in task completion rate \cite{shi2024yell}.
    
\subsubsection{Control}

At the time of writing of this article (December,2024), integration of transformers in control systems remains relatively unexplored, but few selected studies have shown great promise and performance improvements. Decision Transformer (DT)-based control strategies have been shown to generalize easily attaining reasonable performance in previously unseen (zero-shot) control tasks, and outperform other control strategies with 10 shot adaptations, similarly in high perturbation environment DT is shown to generalize better than $H_2$/$H_\infty$. DT has been shown to capture parameter-agnostic structures \cite{zhang2024decision}. TransformerMPC, comprehensively demonstrated use of Transformers to accelerate performance by reducing upto 95.8\% inactive constraints, and substantial time reduction for wheeled bipeds, quadrotors and humanoid robots in the range of 6.8x to 34.9x \cite{zinage2024}. Some researchers have taken a completely different approach using Convolutional Vision Transformer(CvT), "DonkeyNet", as a replacement for MPC, in highly nonlinear scenarios, achieving up to 3x faster times, maintaining consistently low computational time of 20-40 ms while MPC showed significant spikes for complex path points \cite{Hiremath2024}. Architectures like MetaMorph, using transformers to create a universal controller capable of generalizing to 100 different robots, showing strong zero-shot performance and 2-3x more sample efficient during fine-tuning \cite{gupta2022}. 

\section{Challenges and Limitations} 
During the course of this paper, the authors have tried to give a broad understanding about the different ways in which Transformers are being applied in robotics. Despite their potential, several challenges remain, including difficulty in generalization across environments. Foundation Models are being used in various operations from robot manipulation, navigation, to perception. One of the major challenges has been generation of datasets, for training and datasets generated for manipulation may not easily be transferrable to train other subsystems like navigation models for mobile robots. Many researchers are able to create datasets in simulation environment, but scaling and testing same models for real world remains challenging. The largest dataset to date Open X-Embodiment has disproportionate amount of either trajectory or vision data from one particular robot, thus needing corrective measures so as to prevent overfitting of generalist models. There is no guarantee that the zero-short generalization shown by a system in simulation will be recreated to the same extent for real-world deployment. The generalization and human-robot interaction that is demonstrated is often in an indoor setting with manipulation tasks; how to deal with large variations in physical environments, platforms, and range of potential tasks while still maintaining generalization remains a challenge. Recently introduced datasets such as DROID \cite{khazatsky2024droid} provide a gist of the potential improvements that can be achieved with better real-world data. Safety concerns are always heavy, as there are no pre-determined safety guardrails or hard constraints that ensure safety in cases such as zero-shot generalization. An additional challenge is managing the memory requirement during real-time inference and ensuring robustness in adverse weather conditions, such as rain or fog. Future work could involve optimizing attention mechanisms for real-time applications.

\section{Future Directions}
Current SOTA platforms like RT-2 \cite{brohan2023rt2} use cloud deployment and rely on the frequency of responses over the cloud for control, which can limit deployment in applications in areas of low connectivity and online computations on hardware. OpenVLA \cite{kim2024openvla} has demonstrated the use of quantization to reduce the memory footprint of generalist models while maintaining low interference times to avoid latency in the system. Creating generalization for robots under resource-constrained models, and optimization of current computations can enable real-time online computations. Recently CrossFormer \cite{doshi2024scaling}, demonstrated the ability to take in data from any robot embodiment, and the same network weights can control robots of vastly different dynamics, ranging from single and dual arm manipulation systems to quadcopters. Future works with more diverse datasets can provide an opportunity for generalist models capable of working with humans in a collaborative environment. Current generalist policies show variations in performance when deployed across different embodiments. Platforms like DART \cite{park2024dexhub} make low-cost robot manipulation data collection possible using human demonstrations. It is yet to be determined whether or not data collected from platforms like DART can be successfully used to train generalist models, so far no generalist model is using data collected from human demonstrations. The emphasis needs to be on sample-efficient human demonstrations and some research has shown promise for robots to acquire new skills in a few shots \cite{Parakh2024icra}. Other possible directions may include reconciling the performance disparities in simulations with the real world. Some platforms like RealTo \cite{torne2024reconcil} have already shown an excellent increase in the robustness of real-world performance. 

\section{Conclusion}
Transformers have fundamentally transformed NLP with the advent of models like GPT and BERT. Their use in robotics is becoming more prevalent. The study details their role in augmenting human-robot interaction with the use of foundation models such as LLMs. We explore the advancements in zero-shot generalization that have been achieved particularly in the manipulation and controlled environment using various mechanisms including co-training with VLMs. We look over how various augmentations like Vision Transformers and subsequent visuo-linguistic models have provided zero-shot or few-shot generalization capabilities to robots for perception, how transformers integrated with RL have provided the ability to perform tasks over longer horizons, and how predictive abilities of transformers are being used to augment control system performance and reduce computational load. While use of multimodal transformers have given us the ability to integrate various sources, zero-shot generalization in scenarios such as factory floors, search and rescue, caregiving remain untested due to paucity of readily available internet scale robot datasets and whatever availability exists is capital intensive and limited in orders of magnitude lesser when compared to visual or text data. As newer and better architectures are developed and the attention span of robots increases, robots capable of human instruction can meet the demands of real-world environments.

\section{References}
\bibliographystyle{plain}
\bibliography{main}

\end{document}